\documentclass[conference]{IEEEtran}
\IEEEoverridecommandlockouts
\usepackage{cite}
\usepackage{multirow}
\usepackage{amsmath,amssymb,amsfonts}
\usepackage{algorithmic}
\usepackage{graphicx}
\usepackage{textcomp}
\usepackage{xcolor}
\usepackage{hyperref}
\def\BibTeX{{\rm B\kern-.05em{\sc i\kern-.025em b}\kern-.08em
    T\kern-.1667em\lower.7ex\hbox{E}\kern-.125emX}}
\begin{document}

\title{Evidential Temporal-aware Graph-based Social Event Detection via Dempster-Shafer Theory\\
\thanks{Lei Jiang and Hao Peng are the correspongding authors.}
 }

\author{
    \IEEEauthorblockN{Jiaqian Ren$^{1,2}$, Lei Jiang$^{1*}$, Hao Peng$^{3*}$, Zhiwei Liu$^{4}$, Jia Wu$^{5}$, Philip S. Yu$^{6}$}
    \IEEEauthorblockA{$^1$ Institute of Information Engineering, Chinese Academy of Sciences, Beijing, China}
    \IEEEauthorblockA{$^2$ School of Cyber Security, University of Chinese Academy of Sciences, Beijing, China}
    \IEEEauthorblockA{$^3$ School of Cyber Science and Technology, Beihang University, Beijing, China}
    \IEEEauthorblockA{$^4$ Salesforce, San Francisco, USA}
    \IEEEauthorblockA{$^5$ Department of Computing, Univercity of Macquarie, Sydney, Australia}
    \IEEEauthorblockA{$^6$ Department of Computer Science, University of Illinois Chicago, Chicago, USA}
    \IEEEauthorblockA{\{\href{mailto:renjiaqian@iie.ac.cn}{renjiaqian}, \href{mailto:jianglei@iie.ac.cn}{jianglei}\}@iie.ac.cn, \href{mailto:penghao@act.buaa.edu.cn}{penghao}@act.buaa.edu.cn, \href{mailto:zhiweiliu@salesforce.com}{zhiweiliu}@salesforce.com, \href{mailto:jia.wu@mq.edu.au}{jia.wu}@mq.edu.au, \href{mailto:psyu@uic.edu}{psyu}@uic.edu}
    }

\maketitle

\begin{abstract}
The rising popularity of online social network services has attracted lots of research on mining social media data, especially on mining social events. Social event detection, due to its wide applications, has now become a trivial task. 
State-of-the-art approaches exploiting Graph Neural Networks (GNNs) usually follow a two-step strategy: 1) constructing text graphs based on various views (\textit{co-user}, \textit{co-entities} and \textit{co-hashtags}); and 2) learning a unified text representation by a specific GNN model.
Generally, the results heavily rely on the quality of the constructed graphs and the specific message passing scheme. However, existing methods have deficiencies in both aspects: 1) They fail to recognize the noisy information induced by unreliable views. 2) Temporal information which works as a vital indicator of events is neglected in most works.
To this end, we propose ETGNN, a novel Evidential Temporal-aware Graph Neural Network. 
Specifically, we construct view-specific graphs whose nodes are the texts and edges are determined by several types of shared elements respectively. To incorporate temporal information into the message passing scheme, we introduce a novel temporal-aware aggregator which assigns weights to neighbours according to an adaptive time exponential decay formula. Considering the view-specific uncertainty, the representations of all views are converted into mass functions through evidential deep learning (EDL) neural networks, and further combined via Dempster-Shafer theory (DST) to make the final detection. Experimental results on three real-world datasets demonstrate the effectiveness of ETGNN in accuracy, reliability and robustness in social event detection.
\end{abstract}


\section{Introduction}
Nowadays, social media platforms, like Twitter and Facebook, dominate the information digestion for online users. 
Social event detection, which aims to extract and reorganize the media texts into different types of events, can thus benefit greatly in fields like recommendation~\cite{macedo2015context}, disaster risk management~\cite{o2010approaching}, public opinion analysis~\cite{peng2022reinforced} and so on. 
Due to its wide applications, social event detection has been the research hot spot since the last decade~\cite{aggarwal2012event,li2021reinforcement}.

Due to the powerful expressiveness of Graph Neural Networks in capturing both the rich semantics and structural information in social streams, GNN-based methods~\cite{peng2019fine,cao2021knowledge,peng2021streaming,cui2021mvgan,peng2022reinforced} have achieved tremendous success in the social event detection domain.
Generally, the performance of GNN-based methods relies on the quality of the constructed graph and the specific message-passing scheme.
Existing works either simply leverage sharing elements~\cite{cao2021knowledge,cui2021mvgan,peng2022reinforced} or manually design sophisticated meta-paths~\cite{peng2019fine,peng2021streaming} to build graphs and then adopt a specific GNN model to learn a unified representation.
For example, authors in \cite{cao2021knowledge,peng2022reinforced} build a homogeneous message graph in which they add edges between texts sharing common keywords, named entities or users, and then utilize Graph Attention Networks (GAT) to gain comprehensive representations. Even, authors in \cite{peng2019fine,peng2021streaming} design an event meta-schema to build an event-based heterogeneous graph and present a Pairwise Popularity Graph Convolutional Network to make event categorization. 


Despite the effectiveness of aforementioned GNN-based methods, they still suffer from two main limitations.
The first limitation is that they cannot take the connection qualities of graphs under different views (e.g. the different selected common elements or the different designed meta-paths) into consideration.
While in practice, edges between social texts built from varying views usually contain different amounts of information and inevitably introduce some noise or even mistakes. 
It is crucial to dynamically assess the view-specific quality for reliable decision making.
Another limitation is the ignorance of temporal information. 
Since events in the real world always occur at a sudden time and last for a period. 
The time interval between message nodes can work as a powerful indicator of whether two messages belong to the same event. 
However, most recent approaches~\cite{peng2019fine,liu2020event,cao2021knowledge,peng2021streaming,peng2022reinforced} ignore this kind of information. 
Few exceptions~\cite{cui2021mvgan} propose to encode different timestamps into different vectors. 
Failing to capture the time intervals, they are not intuitive enough.
Meanwhile, considering the complicated relation between event proximity and time interval (attention is high in the short-term, and declines sharply in the long-term), it is difficult to design a proper strategy to incorporate temporal information.


To tackle these challenges mentioned above, we propose a novel Evidential Temporal-aware Graph Neural Network, namely ETGNN, to make accurate, reliable and robust social event detection. 
Particularly, three important elements (hashtags, extracted entities and mentioned users) are selected to construct three view-specific message graphs to make the different relations fully explored.
Meanwhile, considering the temporal information, we devise a novel temporal-aware GNN aggregator. 
Inspired by the exponential decay characteristics of event propagation in the real world, we utilize an adaptive time exponential decay formula to make a temporal-aware attention mechanism. Our mechanism assigns different weights to neighbours under the guidance of the time intervals and the corresponding node semantic features. 
After getting the representations of all the views, we apply evidential deep learning neural networks to estimate the uncertainty of each view. 
Those results are further combined via the Dempster-Shafer theory, which is able to consider all the available evidences and give the trusted final decision. 

In experiments, we conduct extensive evaluations on three real-world datasets to validate the effectiveness, reliability and robustness of our model. 
In summary, the contributions of this work can be summarized as follows:
\begin{itemize}
\item We propose a novel Evidential Temporal-aware Graph Neural Network, namely ETGNN, to integrate the uncertainty of view-specific graph quality and temporal information into social event detection.
\item We design a novel temporal-aware GNN aggregator to incorporate temporal information into the message passing scheme. Specifically, we utilize an adaptive time exponential decay formula to make a temporal-aware attention mechanism. 
\item To the best of our knowledge, we are the first to incorporate uncertainty estimation into GNN-based social event detection. With the help of evidential learning and the Dempster-Shafar theory, our model has the ability to dynamically access the quality of each view with evidence and make a trusted decision.
\end{itemize}

\section{Related work}
\subsection{Social Event Detection}
Social event detection aims to detect the occurrences of events and categorize them from the numerous social messages~\cite{aggarwal2012event}. 
Early social event detection methods can be split into content-based methods~\cite{wang2017neural}, and attribute-based methods~\cite{xie2016topicsketch}. 
Recently, with the great success of GNN \cite{kipf2016semi,hamilton2017inductive,peng2021reinforced,peng2021lime}, there has been a move towards GNN-based social event detection~\cite{peng2019fine,cao2021knowledge,peng2021streaming,cui2021mvgan,peng2022reinforced}.
Compared to early studies, GNN-based methods show their superiority in combining content with attributes through message propagation in the graph structure. 
KPGNN \cite{cao2021knowledge} constructs a homogeneous message graph where messages with common elements are connected, and leverages inductive GAT to learn the representations. 
PP-GCN~\cite{peng2019fine} instead adopts weighted meta-paths to consider the different relations well. 
Unfortunately, as meta-paths extracted from complex pre-defined rules inevitably introduce some noise, they are still far from satisfactory. 
Additionally, most existing methods except for \cite{cui2021mvgan} ignore the important temporal information in events.

\subsection{Dempster-Shafer Theory (DST)}
Dempster-Shafer Evidence Theory~\cite{dempster2008upper} is a generalization of the Bayesian theory to subjective probabilities~\cite{dempster1968generalization}. It was later developed into a general framework to give model uncertainty. Due to the flexibility and
effectiveness of DST in modeling uncertainties without prior information, it is widely used in real applications, such as classification~\cite{denoeux2008k} and clustering~\cite{liu2015credal}.




Dempster's rule allows evidence
from different sources to be combined arriving at a degree of beliefs that takes into account all the available evidence. The joint brief, which is the combination of beliefs from different sources is fault-tolerant. Therefore, it is more sufficient and reliable to use the joint belief for reliable decision-making~\cite{su2018research}. 

\begin{figure}
\centering
\includegraphics[width=0.5\textwidth]{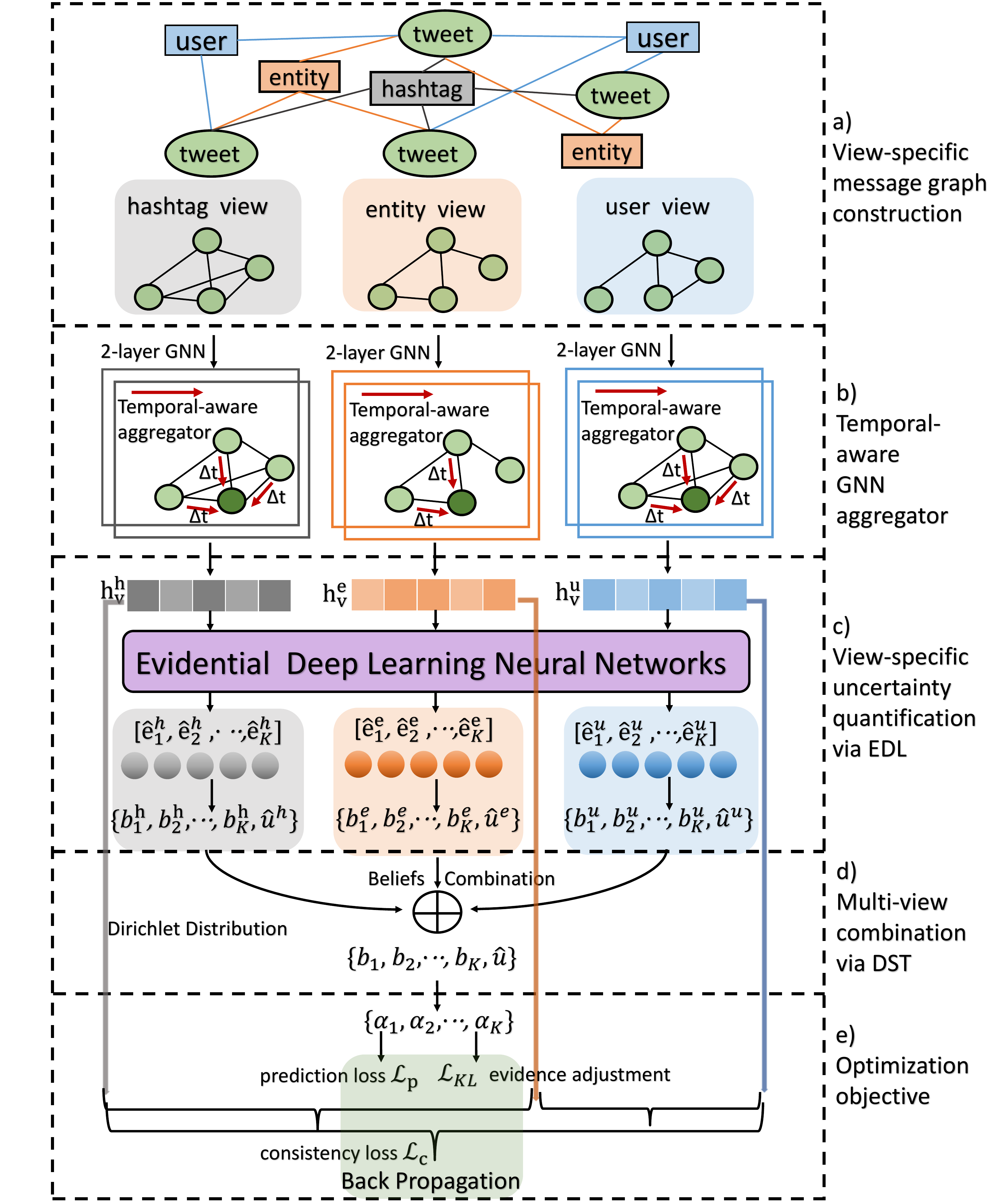}
\caption{The architecture of ETGNN. 
} \label{fig1}
\end{figure}
\section{Methodology}

In this section, we introduce our proposed ETGNN model. As shown in Fig.~\ref{fig1}, the overall framework can be split into five main parts. We start with the construction of view-specific social message graphs in Sec.~\ref{method_gc}. Next, in Sec.~\ref{method_temGNN}, we introduce the novel temporal-aware aggregator in detail. After getting representations of all views, in Sec.~\ref{sec:method_edl}, we utilize
the evidential deep learning neural networks to quantify view-specific uncertainty. Sec.~\ref{method_dst} shows how to combine multi-view beliefs together via Dempster-Shafer theory. Finally, in Sec.~\ref{method_oo}, we demonstrate the optimization objectives including three kinds of losses.

\subsection{View-specific Graph Construction}\label{method_gc}
In the real world, a social event generally refers to influential facts that appear on social networks, which can be highly reflected by the following important elements: involved users, extracted entities, and topics. Since a hashtag often appears as a concise summary of the event, we treat it as the event topic. 

As shown in Fig.~\ref{fig1}(a), we first extract hashtags, entities, and users from tweets, denoted as $h$, $e$ and $u$. 
Next, we treat these three important elements as different views for the construction of tweet graphs, denoted as $G_h=(V,E_h)$, $G_e=(V,E_e)$ and $G_u=(V,E_u)$, where $V$ refers to the set of tweet nodes,
and ${E_h, E_e}$ and ${E_u}$ indicate the edge sets constructed under views from common hashtags, common entities, and common users, respectively. 
We will elaborate more on how to characterize the uncertainty of each view-specific graph in Sec.~\ref{sec:method_edl}.


\subsection{Temporal-aware GNN Aggregator}\label{method_temGNN}

The key to our temporal-aware aggregator is assigning the accurate and reasonable weights to nodes with different node temporal proximity based on their time intervals. 
We refer to the characteristics of the event life cycle in the real world to measure the temporal proximity. Inspired from~\cite{li2012twevent}, the number of tweets associated with an event typically presents a Poisson distribution over time, which indicates that the similarity of the publishing time of any two tweet nodes decreases more rapidly with time. Therefore, we utilize the time exponential decay formula in the following form to measure the temporal approximation: 
\begin{equation}
\rm{sim}(t_i,t_j) = e^{-\lambda\cdot|t_i - t_j|},
\end{equation}
where $t_i$ and $t_j$ are two time points.  $|t_i-t_j|$ is the time interval (days in this work), corresponds to the $\bigtriangleup t$ in Fig.~\ref{fig1}(b). $\lambda$ is the decay rate. Instead of setting a fixed constant $\lambda$, we replace it with a learnable fully connection layer whose input is the representation of the target node. In this way, the decay rate of different events can be learnt adaptively.

We now describe it in a single-layer forward process. Assume that the node $n$ is the target node, and the node representation in layer $l$ is $\mathbf{h}_n^l$. The adaptive temporal-aware GNN aggregator function can be represented as follows:
\begin{equation}
\mathbf{h}^{l+1}_n = \sigma(\sum_{n'\in\mathcal{N}(n)}a_{nn'}\mathbf{W}\mathbf{h}_{n'}^l),
\end{equation}
where $\mathcal{N}(n)$ denotes the neighbour nodes set of the target node $n$. $\textbf{W}$ is the learnable shared weight matrix. $a_{nn'}$ is the corresponding attention weight, which is calculated as follows:
\begin{equation}
a_{nn'} = \frac{e^{-fc(h_n^l)\cdot|t_{n'}-t_{n}|}}{\sum_{n'\in\mathcal{N}(n)}e^{-fc(h_n^l)\cdot|t_{n'}-t_n|}},
\end{equation}
where $fc(\cdot)$ denotes a fully connection layer.

\subsection{View-Specific Uncertainty Measurement with Evidence}\label{sec:method_edl}
As shown in Fig.~\ref{fig1}(c), we directly model the classification uncertainty and high-order probabilities for a prediction in each view from the evidence deep learning perspective. 

To achieve this, with the notion of belief masses (probabilities to the possible sets of class labels) and uncertainty mass (overall uncertainty of the framework) borrowed from Dempster-Shafer theory~\cite{dempster2008upper}, we place a Dirichlet distribution over the estimated posterior probabilities based on the evidence from data~\cite{sensoy2018evidential}.
Formally, for each view $v\in\{h,e,u\}$, suppose there are $K$ mutually exclusive event class labels, we provide a belief mass $b_k^v$ for each singleton event label $k= 1, ..., K$ as well as an
overall uncertainty mass $\hat{u}^v$ of view $v$. These $K + 1$ mass values are all non-negative and sum up to one:
\begin{equation}
\hat{u}^v + \sum_{k=1}^Kb_k^v=1 ,
\end{equation}
where $\hat{u}^v \geq0$ and $b^v_k \geq0$ indicate the overall uncertainty of view $v$ and the probability for the $k$-th class respectively.
As mentioned above, we define the above belief assignment $\mathbf{b}^v = [b^v_1, b^v_2, ..., b^v_K]$ over a Dirichlet distribution $\mathbf{\alpha}^v = [\alpha^v_1, \alpha^v_2, ..., \alpha^v_K]$ with parameters induced from the evidence $\mathbf{\hat{e}}^v = [\hat{e}^v_1, \hat{e}^v_2, ..., \hat{e}^v_K]$ collected from the data (i.e., $\alpha_k^v = \hat{e}_k^v+1$). Noted that the evidence vector can be obtained by an evidential deep learning neural network (i.e. a classical neural network whose $softmax$ layer is replaced with an activation layer). The computation process is as follows:
\begin{equation}\label{equ_u}
b_k^v=\frac{\hat{e}^v_k}{S^v}, \hat{u}^v=\frac{K}{S^v},
\end{equation}\label{equ_bu}where $S^v=\sum_{k=1}^K\hat{e}^v_k+1=\sum_{k=1}^K\alpha_k^v$ is referred to as the Dirichlet strength. In this way, the uncertainty is inversely proportional to the total evidence. 
Meanwhile, different from the output of a standard neural network classifier which is a simple probability assignment, the Dirichlet distribution represents the density of each such probability assignment thus modelling second-order probabilities and uncertainty directly. 


\subsection{Multi-View Combination}\label{method_dst}
As shown in Fig.~\ref{fig1}(d), we utilize the Dempster–Shafer theory to fuse information from different views together.
The combination rule, known as Dempster's rule, strongly emphasizes the agreement between multiple sources and ignores all the conflicting evidence through a normalization factor. Specifically, given two probability mass assignments, $M^h=\{\{b^h_k\}_{k=1}^{K},\hat{u}^h\}$ from the hashtag view and $M^e=\{\{b^e_k\}_{k=1}^{K},\hat{u}^e\}$  from the entity view, the combined joint mass $M=\{\{b_k\}_{k=1}^{K},\hat{u}\}$ is calculated in the following manner:
\begin{equation}
\begin{aligned}
    M = M^h\oplus M^e,\\
    b_{k}=\frac{1}{1-T}\left(b_{k}^{h} b_{k}^{e}+b_{k}^{h} \hat{u}^{e}+b_{k}^{e} \hat{u}^{h}\right), \hat{u}=\frac{1}{1-T} \hat{u}^{h} \hat{u}^{e},
\end{aligned}
\end{equation}
where $T=\sum_{i\neq j}b_i^hb_j^e$
is a measure of the amount of conflict between the two mass sets, and the scale factor $\frac{1}{1-T}$
is used for normalization, which has the effect of completely ignoring conflict and attributing any mass associated with conflict to the null set. 
The operator $\oplus$ is commutative and associative. When the number of views $V$ is larger than $2$ (e.g. we have $3$ views to make event categorization), we can sequentially combine the beliefs from different views with Dempster’s rule of combination as follows:
\begin{equation}
 M = M^1\oplus M^2\oplus...\oplus M^V.
\end{equation}
After the combination of all the views, we obtain the final joint mass $M=\{\{b_k\}_{k=1}^K,\hat{u}\}$. According to Equ.~\ref{equ_u}, the parameters of the Dirichlet distribution are induced as:
\begin{equation}
\alpha_k = b_k \times \frac{K}{\hat{u}} + 1,
\end{equation}
which can be utilized to model the final probability distribution for the class probabilities and the overall uncertainty.

\subsection{Optimization Objectives}\label{method_oo}
Our optimization objective function includes three parts: the prediction error term, the evidence
adjustment term and the consistency constraint.

For the prediction error, we make a simple modification on the traditional cross-entropy loss. The modified loss in essence is the integral of the cross-entropy loss function on the simplex determined by the Dirichlet distribution with the parameter $\mathbf{\alpha}_i$:
\begin{equation}
\begin{aligned}
    \mathcal{L}_{p}=\sum_{i}{\int\left[\sum_{j=1}^{K}-y_{i j} \log \left(p_{i j}\right)\right] \frac{1}{B\left(\mathbf{\alpha}_{i}\right)} \prod_{j=1}^{K} p_{i j}^{\mathbf{\alpha}_{i j}-1} d \mathbf{p}_{i}}
    ,
\end{aligned}
\end{equation}
where $\mathbf{y}_i$ is the true class distribution. $\mathbf{p}_i$ is the class assignment probabilities on a simplex and $\it{B}(\cdot)$ is the multinomial beta function.

During training, we expect the total evidence to shrink to zero if it cannot be correctly classified. We achieve this by incorporating a Kullback-Leibler (KL) divergence as the evidence
adjustment term. The loss is:
\begin{equation}
    \begin{array}{l}
\mathcal{L}_{KL}=\sum_{i}\text{KL}\left[D\left(\mathbf{p}_{i} \mid \tilde{\mathbf{\alpha}}_{i}\right) \| D\left(\mathbf{p}_{i} \mid \mathbf{1}\right)\right]
,
\end{array}
\end{equation}
where $\tilde{\mathbf{\alpha}}_{i}=\mathbf{y}_{i}+\left(1-\mathbf{y}_{i}\right) \odot \mathbf{\alpha}_{i}$ is the Dirichlet parameters after removal of the non-misleading evidence from predicted parameters. $\it{D}(\cdot|\cdot)$ denotes the multinomial opinions formed by the Dirichlet parameter.

Meanwhile, considering Dempster's rule fails to give satisfactory results when high degree of conflict between different views, we additionally add a consistency constraint in the representations of tweets obtained by the three views. Suppose the normalized embeddings over a batch of training samples from the hashtag, entity and user view are denoted as $\mathbf{H}^h_{nor}$, $\mathbf{H}^e_{nor}$, and $\mathbf{H}^u_{nor}$, the similarity of nodes for a specific view in a batch is as follows:
\begin{equation}
\begin{aligned}
    \mathbf{C}^{v}=\mathbf{H}^v_{nor} \cdot (\mathbf{H}^v_{nor})^{T}, v\in\{h,e,u\}
\end{aligned}
\end{equation}
The consistency implies that the three similarity matrices should
be similar, which gives rise to the following constraint:
\begin{equation}
    \mathcal{L}_{c}=\left\|\mathbf{C}^{h}-\mathbf{C}^{e}\right\|_{F}^{2}+
    \left\|\mathbf{C}^{h}-\mathbf{C}^{u}\right\|_{F}^{2}+
    \left\|\mathbf{C}^{e}-\mathbf{C}^{u}\right\|_{F}^{2}.
\end{equation}

Accordingly, the overall optimization objective function is as follows:
\begin{equation}
    \mathcal{L}_{total}= \mathcal{L}_{p} +\lambda_e\mathcal{L}_{KL}+\lambda_c\mathcal{L}_{c},
\end{equation}
where $\lambda_e$ controls the weight of evidence regularization and encodes the tolerance to accept misleading samples. $\lambda_c$ controls the weight of conflict penalty.

\begin{table*}[h]
\caption{Performance Comparison with ETGNN and Baselines.}\label{tab_result}
\centering
\renewcommand\arraystretch{1.05}
\setlength{\tabcolsep}{7mm}
\begin{tabular}{l|l|l|l|l|l|l}
\hline\hline
\multirow{2}{*}{Methods} & \multicolumn{2}{c|}{Events2012}  & \multicolumn{2}{c|}{Kawarith}  & \multicolumn{2}{c}{CrisisLexT26}\\
\cline{2-7}
\multirow{2}{*}{} &ACC &F1 & ACC &F1 &ACC &F1\\
\hline
TwitterLDA &0.3796   &0.1376 &0.5281 &0.4484 &0.5275 &0.5212\\
JETS&0.5943   &0.5614  &0.6579 &0.6444 &0.7154 &0.7008\\
Word2Vec&0.4484  &0.4300  &0.3980  &0.3565 &0.6484  &0.6464\\
BERT&0.7768   &0.6420  &0.9140 &0.9083 &0.9242 &0.9240\\
PP-GCN&0.6120  &0.6171 &0.8776 &0.8671 &0.8113 &0.8057\\
KPGNN &0.7876  &0.6899 &0.9360 &0.9309 &0.9643 &0.9646\\
MVGAN&0.7800   &0.7358 &0.9432 &0.9410 &0.9720 &0.9680\\
\hline
ETGNN&\textbf{0.8480}  &\textbf{0.7565} &\textbf{0.9636} &\textbf{0.9610} &\textbf{0.9792} &\textbf{0.9794}\\
\hline\hline
\end{tabular}
\end{table*}
\section{Evaluation}

\begin{figure*}
\centering
\includegraphics[width=0.82\textwidth]{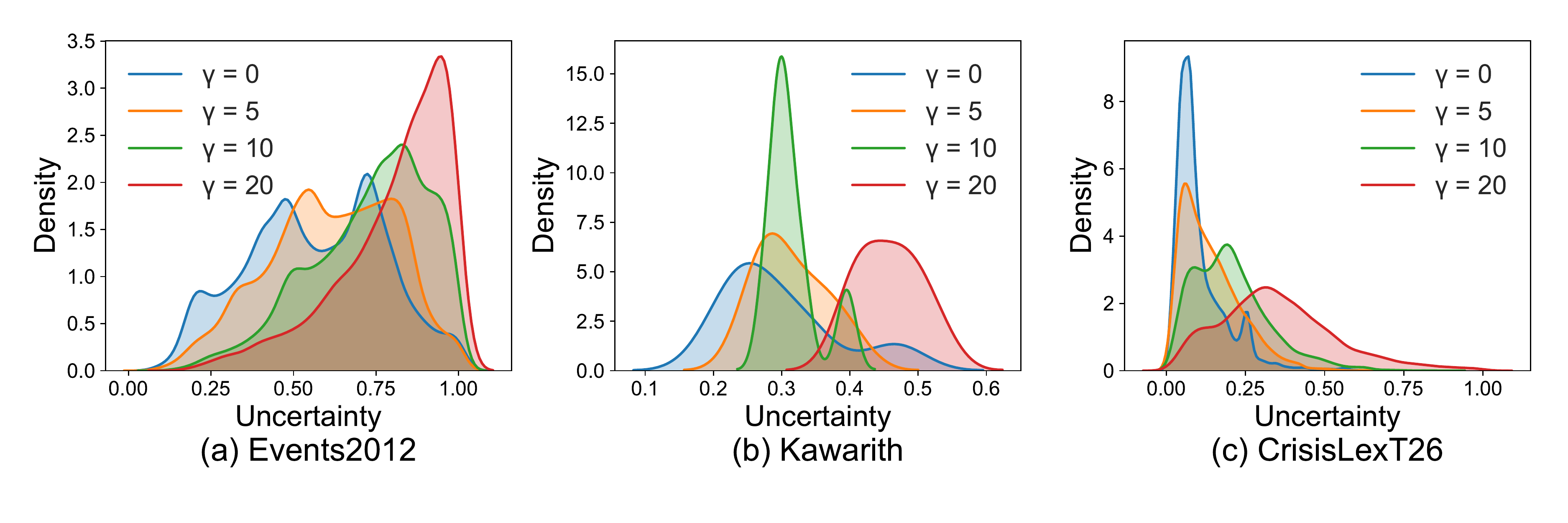}
\caption{Uncertainty with data noise. 
} \label{fig_noise}
\end{figure*}

\subsection{Datasets}
We conduct experiments on three large-scale, publicly available datasets, the English Twitter dataset Events2012~\cite{mcminn2013building}, 
the Arabic crisis event dataset Kawarith~\cite{alharbi2021kawarith} and the multilingual dataset CrisisLexT26~\cite{olteanu2015expect}. After filtering out repeated and irretrievable tweets, the statistics of three datasets are as follows:
(1) Events2012: This English Twitter dataset contains 68,841 manually labeled tweets related to 503 event classes, spread throughout four
weeks. 
(2) Kawarith: There are 9,070 labeled tweets
relating to 7 crisis-class events over different periods.
(3) CrisisLexT26: The dataset used in our research is a small subset of CrisisLexT26.
7 crisis events in English with 6,733 labeled tweets between 2012 and 2013 are selected.

\subsection{Baselines and Experimental Settings}
\subsubsection{Baselines}
We compare ETGNN to both non-GNN-based methods and GNN-based methods. For the former, the baselines are: 
(1) \textbf{TwitterLDA}~\cite{zhao2011comparing}, which is the first proposed topic model for Tweet data; (2) \textbf{JETS}~\cite{wang2017neural}, which is a deep learning method for short text representation by designing a neural model to joint event detection and summary; (3) \textbf{Word2Vec}~\cite{mikolov2013efficient}, which uses the average of the pre-trained embeddings of all words in the message as its representation; (4) \textbf{BERT}~\cite{devlin2018bert}, which combines the sentences embeddings of BERT with a conventional two-layer neural network classifier.
For GNN-based methods, we select
(5) \textbf{PP-GCN}~\cite{peng2019fine}, an offline fine-grained social event
detection method based on GCN; (6) \textbf{KPGNN}~\cite{cao2021knowledge}, which leverages GAT to efficiently detect events; (7) \textbf{MVGAN}~\cite{cui2021mvgan}, which utilizes GAT to learn representations of tweets from both semantic and temporal views.


\begin{figure*}
\centering
\includegraphics[width=0.82\textwidth]{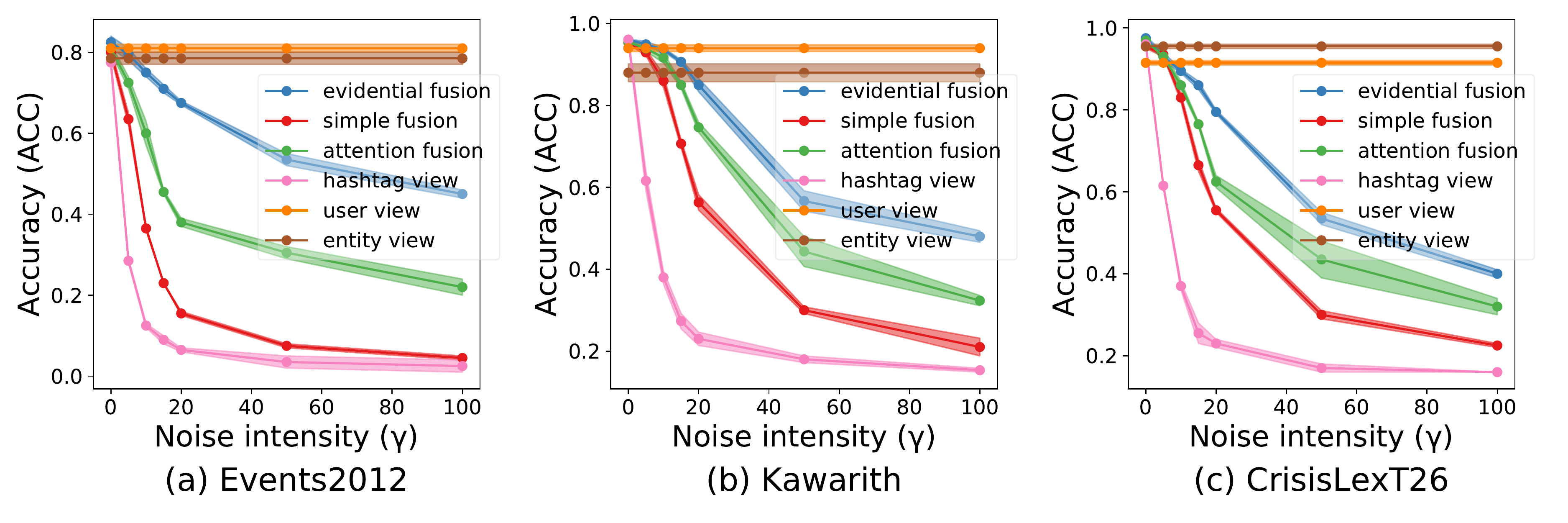}
\caption{Uncertainty in improving model robustness. 
} \label{fig_improve}
\end{figure*}

\subsubsection{Experimental Settings}
In ETGNN, we set the batch size to 2000, the total number of GNN layers to 2, the first and second layer embedding dimension to 512 and 256, respectively. As for the EDL part, we use a two-layer neural network with the $softmax$ layer replaced by an activation layer ReLU to ascertain
non-negative evidence. The hidden layer dimension is 128. As for the optimization objective, we set $\lambda_e$ to 1 and $\lambda_c$ to 5.
We set the learning rate to 0.001, the optimizer to Adam, and
the training epochs to 100. 

\subsection{Evaluation Results}
This subsection compares ETGNN to the baselines. 
We split 80\%, 10\% and 10\% of the data as the training set, validation set and the test set, respectively. To avoid the one-time occasionality, we perform 10 tests for all algorithms on all three datasets, and record the average values.

Table~\ref{tab_result} summarizes the results.
ETGNN dominates all baseline algorithms in all evaluation metrics by large margins. That is because ETGNN integrates two important guidance information: 1) the quality estimation of constructed graphs from different views and 2) the vital temporal information. Generally, compared to methods simply relying on measuring the distributions of words (TwitterLDA) or text embeddings (Word2vec, BERT), GNN-based algorithms (PP-GCN, KPGNN, MVGAN) get better results. This emphasizes the importance of GNN's capacity in leveraging the underlying graph structure and conceptual semantics simultaneously.
The MVGAN outperforms the KPGNN method that also uses the GAN framework (4.59\% improvements in F1 score on Events2012). Because MVGAN additionally integrates the temporal information into the final representations. Our ETGNN, also considering temporal information, performs even better than MVGAN. The performance is improved by about 6\% in Events2012. This improvement can be attributed to the uncertainty-based fusion strategy. With the help of EDL and DST, our ETGNN is aware of the view-specific noise
and thus achieves impressive results on all datasets. 

\subsection{Uncertainty Analysis}

\subsubsection{Ability of capturing uncertainty} 

We investigate the relation between uncertainty and data noise. 
There are three views in our model. We add noise to data in the hashtag view.
Specifically, we generate noise vectors that are sampled from Gaussian distribution $N(0,I)$ and add these
noise vectors multiplied with intensity $\gamma$ to
pollute the learnt representations.
The results are shown in Fig.~\ref{fig_noise}. As the noise intensity increases, the uncertainty of noisy samples also increases. This
demonstrates that the estimated uncertainty is closely related
to data quality, which further validates that our proposed EVGNN does capture the data quality.



\subsubsection{Uncertainty in improving model performance}
To verify the superiority of the evidential fusion strategy in ETGNN, we compare it with other two common fusion strategies: 1) the simple fusion strategy, which simply uses the sum of decisions from all the views as the final decision; 2) the attention fusion strategy, which leverages the attention mechanism to fuse all the views together. 
    We conduct experiments in the polluted datasets adding varying degrees of noise to the hashtag view. The results are shown in Fig.~\ref{fig_improve}. Obviously, with the increase of the noise intensity, classification performance on the noisy view (i.e. the hashtag view) decreases rapidly while the performance of the other two views keeps stable. Meanwhile, the results of the other two fusion strategies drop more rapidly compared to the evidential fusion strategy in our model. In other words, with the help of uncertainty, ETGNN is more robust on noisy data.

\section{Conclusion}

In this work, we propose a novel Evidential Temporal-aware Graph Neural Network model to do social event detection task. Specifically, we consider the uncertainty of constructed graphs under different views based
on Evidential Deep Learning and Dempster-Shafer theory. The estimated uncertainty provides guidance for multi-view integration, which leads to more robust and interpretable classification. Meanwhile, we also devise a novel temporal-aware GNN aggregator to incorporate temporal information.
Extensive experiments show the effectiveness and the
superiority of our model compared to methods without taking uncertainty and temporal information into consideration. 

\section*{Acknowledgment}
The authors of this paper were supported by the National Key R\&D Program of China through grant 2021YFB1714800, Natural Science Foundation of Beijing Municipality through grant 4222030 and NSFC through grant 62002007. 
We also thank CAAI-Huawei MindSpore Open Fund and Huawei MindSpore platform for providing the computing infrastructure.

\bibliographystyle{IEEEtran}
\bibliography{IEEEabrv,mybibliography}


\end{document}